\documentclass[smallextended]{svjour3}       
\usepackage{eso-pic}
\newcommand\AtPageUpperMyright[1]{\AtPageUpperLeft{%
 \put(\LenToUnit{0.5\paperwidth},\LenToUnit{-1cm}){%
     \parbox{0.5\textwidth}{\raggedleft\fontsize{9}{11}\selectfont #1}}%
 }}%
\newcommand{\conf}[1]{%
\AddToShipoutPictureBG*{%
\AtPageUpperMyright{#1}
}
}
\usepackage{graphicx,amsmath,geometry,subcaption,hyperref}
 \usepackage[ruled,vlined]{algorithm2e}
\usepackage{algorithmic,latexsym}
\graphicspath{{./figures/}}
\begin{document}
\title{Collaborative Tracking and Capture of Aerial Object using UAVs}
\author{ Lima Agnel Tony$^\dagger$ \and   Shuvrangshu Jana$^\dagger$ \and Varun V P$^*$  \and Vidyadhara B V$^\dagger$ \and Mohitvishnu S Gadde$^\dagger$   \and Abhishek Kashyap$^\dagger$ \and Rahul Ravichandran$^\dagger$ \and Debasish Ghose$^\dagger$}
\institute{ $^\dagger$Guidance Control and Decision Systems Laboratory \\
	Department of Aerospace Engineering\\
	Indian institute of Science\\
	Bangalore-12, India\\
	$^*$Robert Bosch Center for Cyber Physical Systems\\
	Indian Institute of Science\\
	Bangalore-12, India
}
\date{}
\maketitle
\conf{MBZIRC Symposium\\ADNEC, Abu Dhabi, 26-27 February 2020}
\vspace{-1.5cm}
\begin{abstract}
This work details the problem of aerial target capture using multiple UAVs. This problem is motivated from the challenge 1 of Mohammed Bin Zayed International Robotic Challenge 2020. The UAVs utilise visual feedback to autonomously detect target, approach it and capture without disturbing the vehicle which carries the target. Multi-UAV collaboration improves the efficiency of the system and increases the chance of capturing the ball robustly in short span of time. In this paper, the proposed architecture is validated through simulation in ROS-Gazebo environment  and is further implemented on hardware.
\end{abstract}
\section{Introduction}\label{S0}
Aerial target capture involves accurate measurement of the target parameters and is often constrained by various physical limitations of the agents involved in the operation. A robust autonomous architecture demands a comprehensive software and hardware design. The problem could be subdivided into various components like vision to detect the target, guidance and estimation to appropriately approach the target, controller to avoid deviations from the guidance and manipulator module to capture the target. Thus, the complete solution for a vision based aerial target capture is multi-disciplinary in itself. There is considerable literature on aerial interception of static and dynamic targets. Stereo vision based interception of static or moving targets is discussed in \cite{rf1}. Interception and surveillance for counter UAV applications is detailed in \cite{rf2}. A vision based UAV interception method developed on ROS is detailed in \cite{rf3}. A fuzzy logic controller based target tracking and interception is discussed in \cite{rf4} while \cite{rf5} details dynamic ground target neutralization using nonlinear control methods.A PN guidance based strategy to follow and neutralize a combatant UAV is explained in \cite{rf6}. Vision based autonomous path planning for interception is described in \cite{rf7} while \cite{rf8} details the motion planning strategies for intercepting radio active sources. Anti-UAV solution in a GPS denied environment using machine learning is described in \cite{rf9}. A stereo vision based detection and neutralization of non-cooperative drones while localizing the intruder is discussed in \cite{rf9a}.

In this work, we address the problem of aerial interception. Multiple UAVs are involved in achieving this. This problem is different from the regular interception problems in the sense that the target carrying vehicle need not be damaged but the target alone needs to be intercepted. This brings in the safety aspects associated with interception as well. This paper lays out the different aspects in terms of hardware and software, required to achieve the safe target interception. Vision algorithm for target detection, aerial manipulator design and prototyping are also detailed in this work. 

This paper is organised as: Section \ref{S1} defines the problem while Section \ref{S2} formulates the solution detailing the multiple aspects that constitute the same. Section \ref{S3} details the hardware implementation details and Section \ref{S4} gives the simulation and real time results obtained. Section \ref{S5} concludes the work.
\section{Problem description}\label{S1}
\begin{figure}
	\centering
	\includegraphics[scale=0.3]{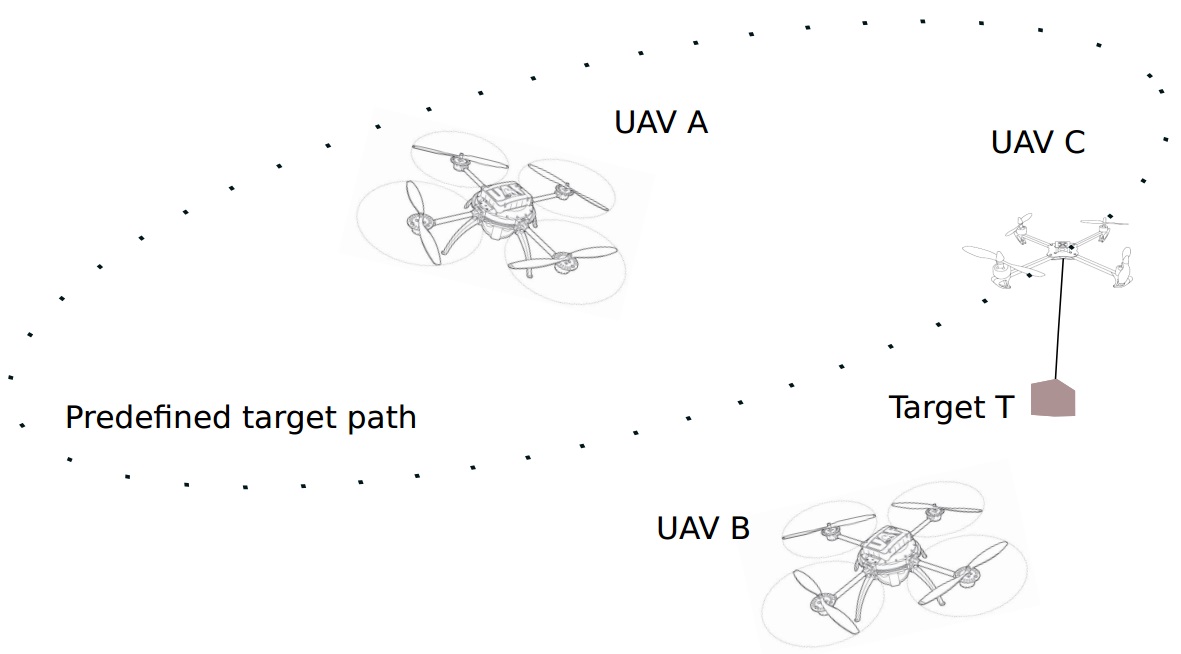}
	\caption{Target bearing UAV moves in some pattern. Own UAVs should capture the target which is suspended from the UAV.}
	\label{f1}
\end{figure}

To start with, the UAVs are considered to be drones and the target to be captured is suspended below another drone. The UAV with the suspended target moves at a certain speed in a certain pattern. The objective of the own UAVs would be to collaborate among each other to capture the target without dashing on to the target bearing UAV. The target is attached to the UAV such that the own UAVs require to apply some detachment force to release the target. The problem could be represented as in Fig. \ref{f1}. UAV A and UAV B are own UAVs while UAV C is the target bearing UAV. The UAVs A and B should coordinate to detach the target T from UAV C. For practical applications, this could be considered as package hand over from one UAV to another for long distance courier delivery, the hand over owing to the endurance limits of vehicles.
\section{Formulation}\label{S2}
Breaking down the problem, following as the essential components for achieving a successful capture of target using vision.
\subsection{Target detection}
\begin{figure}
    \centering
\begin{subfigure}{0.33\textwidth}
\includegraphics[scale=0.45]{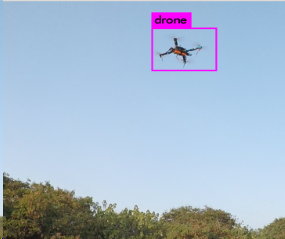} 
\caption{}
\end{subfigure}
\begin{subfigure}{0.3\textwidth}
\includegraphics[scale=0.185]{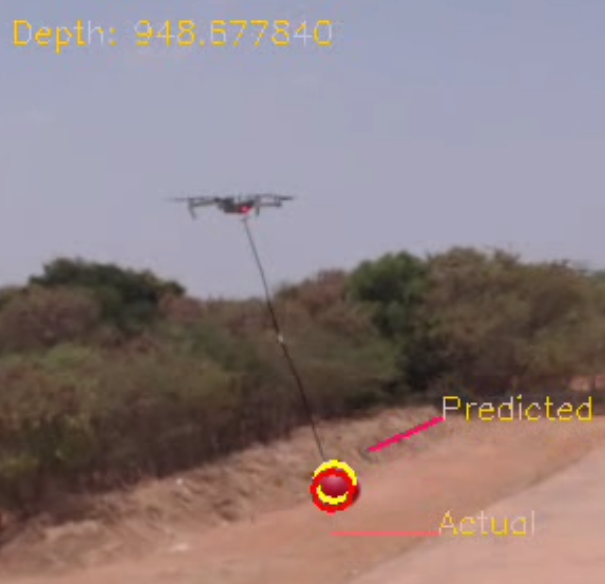} 
\caption{}
\end{subfigure}
\caption{(a) Drone detection using Tiny YOLO v3 (b) Ball detection using segmentation}
\label{f2}
\end{figure}

For demonstration purposes, we have considered a red ball as the target. This is of diameter 18 cm and is suspended using a rod of length 1.5 m from the UAV. As the vehicle moves, the ball sways resulting in a pendulum like motion. The requirements from the vision module include detection of drone carrying the target and the target itself. An object detector based on shape and color is utilized to detect the red ball attached to the target drone. A lighter version of YOLO – Tiny YOLO v3\cite{rf10} is trained to identify drones and is computationally much less expensive on deployment. The data sets used for drone detection are from Yuen University, METU and USC. The data sets have a wide variety of drones at different scales, with a total of 58,985 images. Detecting the drone will give an approximate area for detecting the ball hanging from it, which will improve tracking of the ball. This is especially needed when the ball is far away and appears very small. It also helps resolve ambiguities in case multiple balls or similar objects are found in the scene. An instance of detection of drone and ball is shown in Fig. \ref{f2}.
The ball detection algorithm is enhanced by integrating a Kalman Filter tracker to improve the accuracy at larger distances. The tracker is initialized from a nearby distance. The algorithm is also improved to account for scenarios where the ball might be out of the field of view of the drone (i.e., out of the image) but may reappear later. The distance of the ball is estimated using simple relations as the size of the target is considered known.
\subsection{Image based guidance}
\begin{figure}
	\centering
	\includegraphics[scale=0.25]{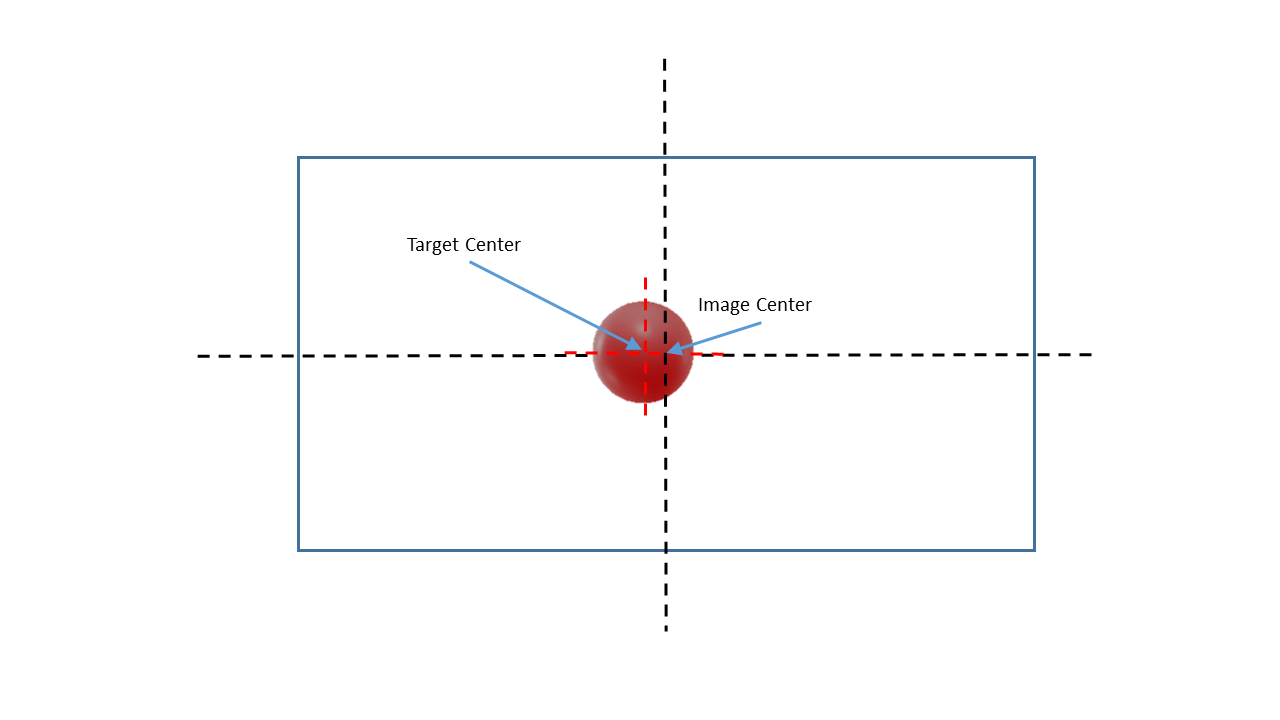}
	\caption{The ball in camera field of view (FOV). The control makes sure that the ball is always within an area of interest}
	\label{f3}
\end{figure}
The grabber drone is subjected to guidance commands in terms of velocity and yaw rate  based on the pixel-coordinates of the ball in the camera frame.  
The grabber drone is fitted with monocular camera. The visual feedback includes the center of the ball as well as the estimated distance of the ball from the center of the camera (Fig. \ref{f3}).
Let ($x_i$, $y_i$) be the coordinates of the target in the image frame and $w$ and $h$ be the width and height of the detected target in the image frame in pixels. Let $W$ and $H$ denote the width and height of the image frame in pixels. In camera frame yaw rate ($\dot{\psi}_{des}$) and height ($\dot{z}_{des}$) control commands are used as inputs to keep the target as close as possible to the image centre.  As a monocular camera is used, the change in the image size is used to determine the change in the range ($r$). Velocity command in the  camera frame $x$ direction ($\dot{x_{des}}$)  is used to reduce the range error. The UAV camera is fixed to the body frame with the optical axes aligned towards the $x$ axis. The commanded velocity in the camera frame is as follows: 
\begin{eqnarray}
	\dot{\psi}_{des}&=& kp_{\psi}(\frac{W}{2}-x_i)+kd_{\psi}(-\dot{x}_i)\\
	\dot{z}_{des}&=& kp_{z}(\frac{H}{2}-y_i)+kd_{z}(-\dot{y}_i)\\
	\dot{x}_{des}&=& kp_{r}(r_{des}-r)+kd_{r}(-\dot{r})
\end{eqnarray}
where, $kp_{\psi}$, $ kd_{\psi}$, $kp_{z}$, $kd_{z}$, $kp_{r}$, $kd_{r}$ are  appropriate gains selected based on the  dynamics of the manipulator bearing  drone.
The desired velocity in the camera frame transformed to vehicle frame and finally commanded to interceptor drone. When the distance between the target UAV and own UAV is within a certain limit, the target tracking switches to the red ball. Kalman filter is implemented to improve the estimation of the image coordinates. 
\subsection{Manipulator design}
For capture of any aerial object, a wide variety of manipulators could be implemented. Each design has its own advantages and disadvantages. The degrees of freedom is one main factor which could affect the performance of the complete system. In this work, we propose a passive basket type manipulator. First of all, the energy budget of the system is optimised by avoiding an active manipulator. Besides that, for any active manipulation mechanism, exact timing is essential for its engagement as its primary objective is to grab a moving ball. Any mistiming would lead to failure of capture of the target even though every other module of the systems works properly. Another aspect in this regard is the fail safe associated with the system. Any jamming in the moving parts of the manipulator would result in the failure of the entire mission. 
\begin{figure}
    \centering
\begin{subfigure}{0.33\textwidth}
\includegraphics[scale=0.25]{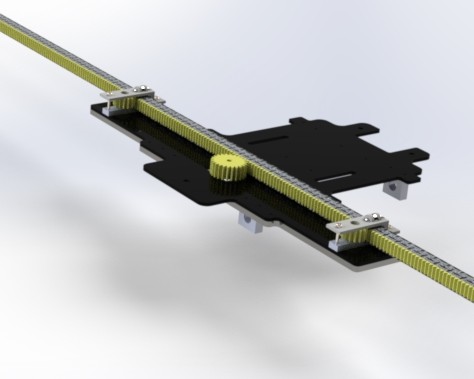} 
\caption{}
\end{subfigure}
\begin{subfigure}{0.3\textwidth}
\includegraphics[scale=0.285]{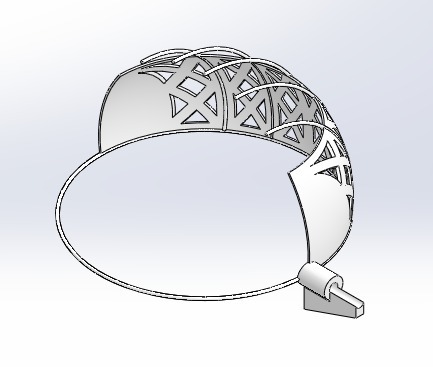} 
\caption{}
\end{subfigure}
\begin{subfigure}{0.3\textwidth}
\includegraphics[scale=0.175]{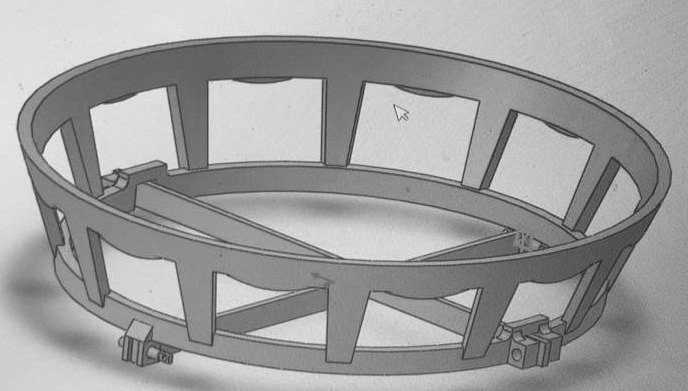} 
\caption{}
\end{subfigure}
\caption{(a) Model of rack and pinion mechanism with the mount  (b) Concept of design of the end effector (c) Target collector }
\label{f4}
\end{figure}
The manipulator arm extends out using a rack and pinion model, as in Fig. \ref{f4}(a). The end effector has a design, as shown in Fig. \ref{f4}(b). The design consists of a ring to which a set of forward facing claws are attached. The purpose of the claws is to detach the target ball.
\subsection{Grab detection}
Upon detaching the target, the UAV proceeds to safe landing. The transition of vehicle to its next task depends upon detecting if the ball has been grabbed or not. For this, a grab detector needs to be suitably mounted. The end effector is attached with a grab detector, concept of which is shown in Fig. \ref{f4}(c). Fig. \ref{f4} shows the entire assembly. The end effector shows a modified claw type passive gripper and a ball collecting base. As seen in Fig. \ref{f4}, proper proximity switches could be attached to the base mount which facilitates detection. Grab detection marks the end of mission.
\subsection{Multi drone collaboration for target capture}
While a single drone could be used to capture a moving target, there are cases in which a single drone brings the efficiency of the system down. One such situation would be the missing of the target. If the target is missed while the UAV tries to grab it, in order to find the target again, the UAV have to execute exploration again. To avoid that, a tracker drone is employed. The job of the tracker is to keep the target ball in FOV once it's detected. This not only helps in avoiding unnecessary search but also in better estimation of the target. 

In this approach, one of the own UAVs take off and start exploration. Once the ball is detected, this information is communicated to the second UAV which carries the manipulation mechanism. The communicated information include the location of the detected ball in world frame. The own UAV moves towards this point. Once the ball is in FOV, the UAV switches on to vision based guidance. A camera sensor is fixed inside the end effector of the manipulator to aid detection. The controller forces the UAV to align the manipulator camera center to the target ball. The UAV proceeds to capture the ball. Once captured, the grab detection mechanism goes high resulting in the end of mission. The tracker drone also reaches end of mission only after the capture of target.
\section{Hardware Implementation}\label{S3}
Target capture using UAV collaboration is implemented on realistic system. Following are the details of the hardware used in this mission.
\subsection{Hardware and sensors used}
\begin{figure}
    \centering
    \begin{subfigure}{0.45\textwidth}
\includegraphics[scale=0.05]{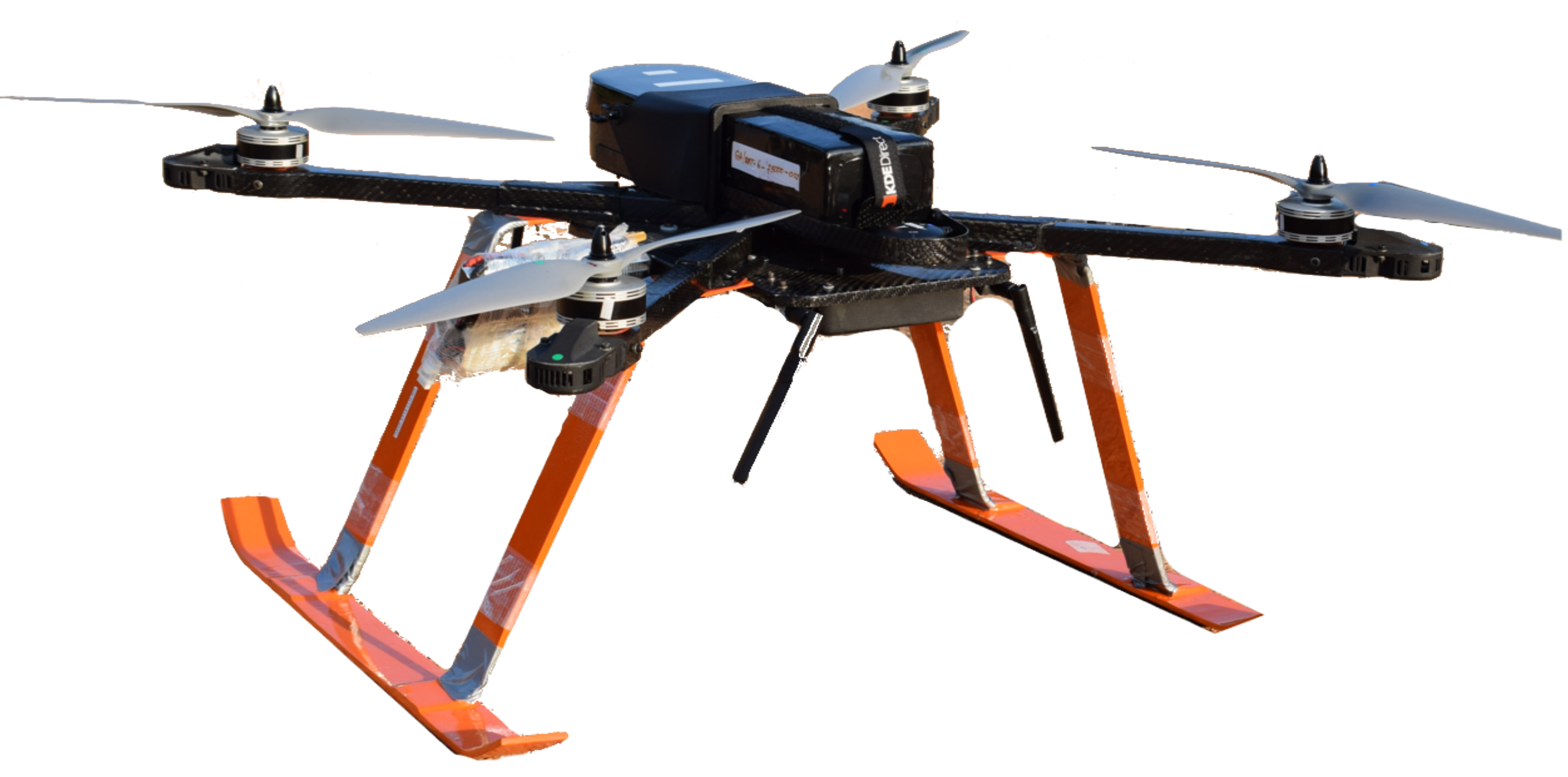} 
\caption{}
\end{subfigure}
\begin{subfigure}{0.4\textwidth}
\includegraphics[scale=0.15]{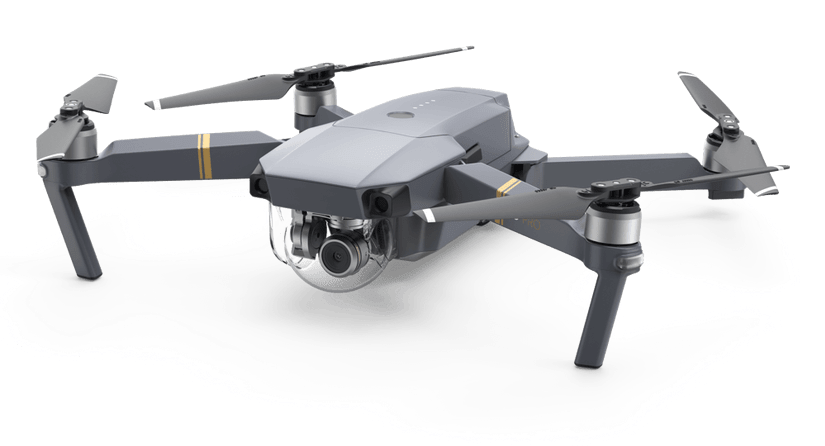} 
\caption{}
\end{subfigure}
    \caption{(a) GA3 series quadrotor platform used as own vehicle (b) DJI Mavic pro platinum as target drone}
    \label{f6}
\end{figure}

Custom designed vehicles are used as the test platform to evaluate the performance of the algorithms being used for the accomplishment of the task. Fig. \ref{f6}(a) is the custom designed quadcopter GA3 series drone. 
This platform is designed keeping mind the thrust to weight ratio requirements for the payload to offer easy integration and customisation options for various sensor suites. GA3 is equipped with an open source Pixhawk 4 autopilot running in Arducopter 3.7.2 firmware and NVIDIA Jetson TX2 with Auvidea J130 carrier board is the onboard computer on the GA3 drone. The NVIDIA Jetson TX2 is equipped with dual-Core NVIDIA Denver 1.5 64-Bit CPU and Quad-Core ARM® Cortex®-A57 MPCore processor with 8 GB 128-bit LPDDR4 of RAM. The computer has an onboard 32 GB eMMC which enhances the data transmission. The target drone is DJI Mavic Pro Platinum (Fig. \ref{f6}(b)).

\begin{figure}
    \centering
        \begin{subfigure}{0.32\textwidth}
\includegraphics[scale=0.095]{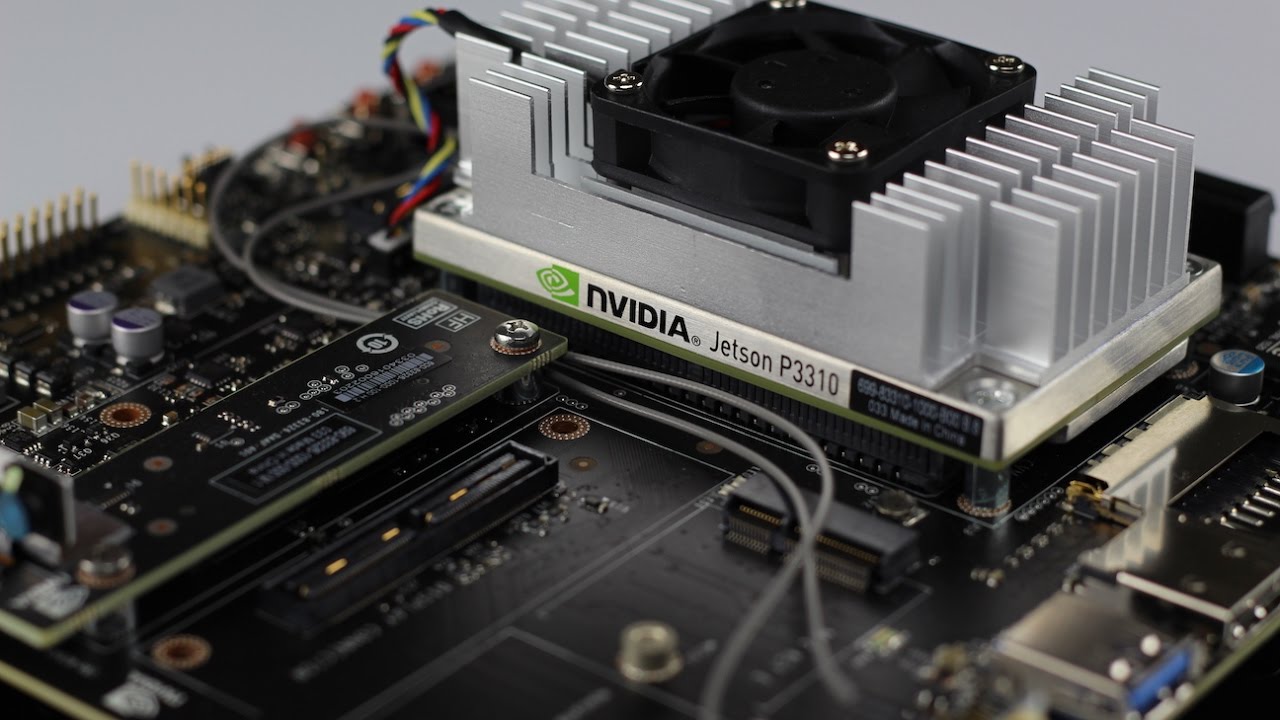} 
\caption{}
\end{subfigure}
    \begin{subfigure}{0.3\textwidth}
\includegraphics[scale=0.125]{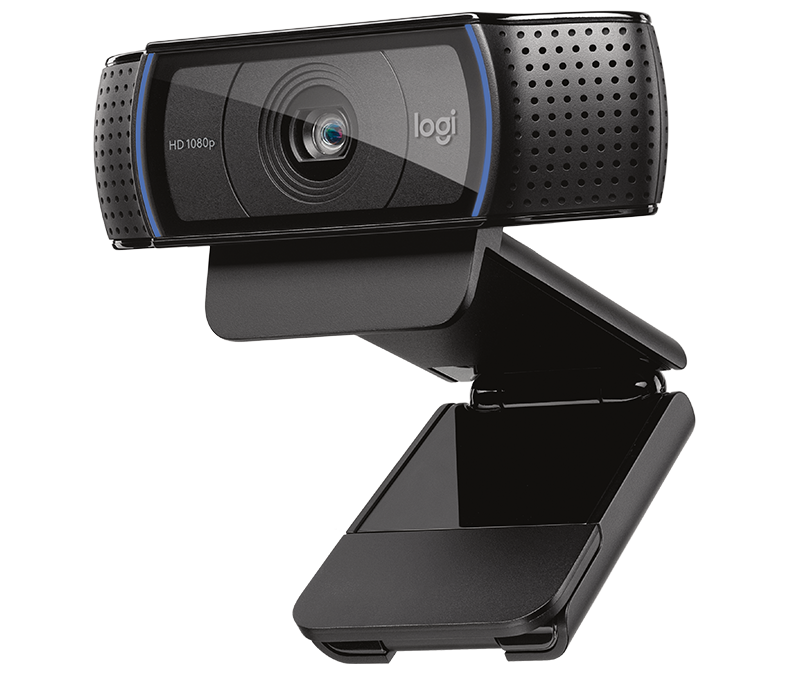} 
\caption{}
\end{subfigure}
\begin{subfigure}{0.33\textwidth}
\includegraphics[scale=0.25]{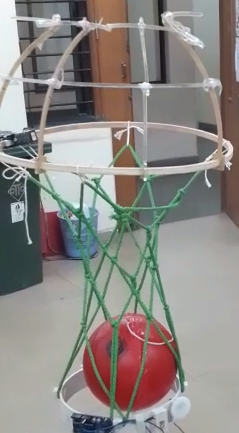} 
\caption{}
\end{subfigure}
    \caption{(a) Nvidia Jetson TX2 (b) Logitech c920 used for target detection and tracking (c) Manipulation mechanism prototyped with the grab detector}
    \label{f7}
\end{figure}

All the processing is done on the onboard NVIDIA Jetson TX2 (Fig. \ref{f7}(a)). The vision stack described in this paper uses an inexpensive out of the shelf Logitech C920 web camera, which is made to run at 480p mode at 30 fps which is mounted on the body frame as well as the manipulator end effector of the vehicle (Fig. \ref{f7}(b)).  The vehicle is also equipped with the state of the art lidar. A high speed (400 Hz) and long range Lightware LW20/C lidar is integrated to the UART port of the Pixhawk 4 autopilot for the accurate measurement of drone's altitude. MAVROS package is used for communicating with the vehicle's autopilot. The computed velocity commands are sent to the MAVROS which then send the required commands to the Pixhawk via Mavlink. The controller node runs at 20 Hz. The manipulator arm is prototyped using carbon fibre tubes for arms. The end effector is fabricated using acrylic sheets, wood and 3D printed parts (Fig. \ref{f7}(c)). the manipulator camera discussed earlier, is placed inside the passive mechanism .
\section{Results}\label{S4}
\begin{figure}
    \centering
    \begin{subfigure}{0.45\columnwidth}
    	\centering
    	\includegraphics[scale=0.235]{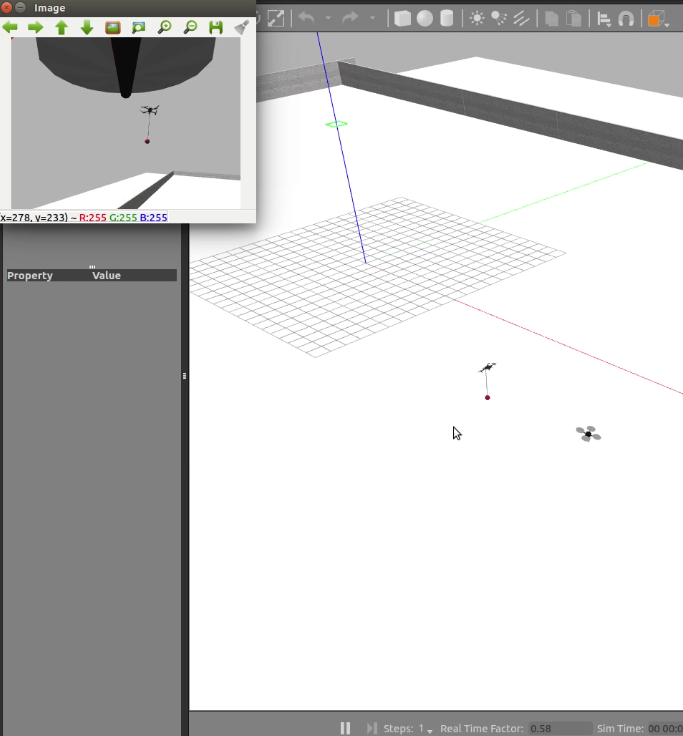}
    	\caption{}
    \end{subfigure}
    \begin{subfigure}{0.45\columnwidth}
	\centering
	\includegraphics[scale=0.25]{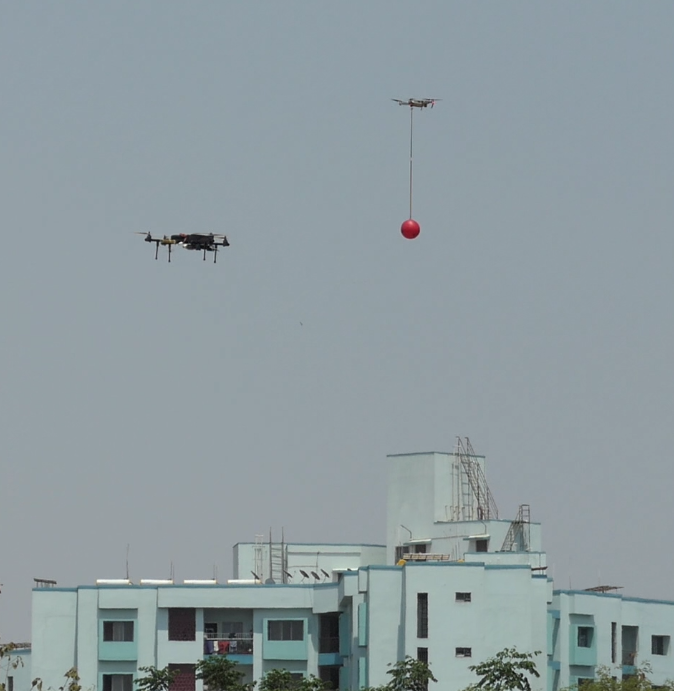}
	\caption{}
\end{subfigure}    
      \caption{(a) UAV target tracking in Gazebo environment and (b) in realistic environment}
    \label{f8}
\end{figure}

\begin{figure}
    \centering
        \begin{subfigure}{0.45\columnwidth}
    	\centering
    	 \includegraphics[scale=0.175]{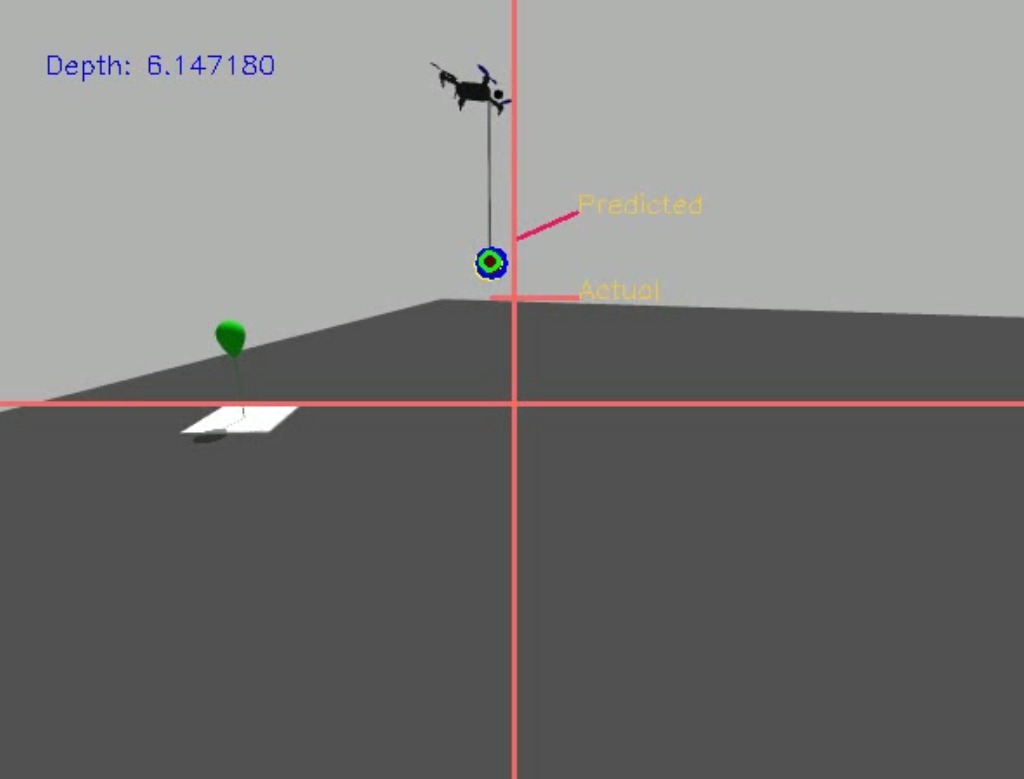}
    	\caption{}
    \end{subfigure}
    \begin{subfigure}{0.45\columnwidth}
	\centering
	  \includegraphics[scale=0.175]{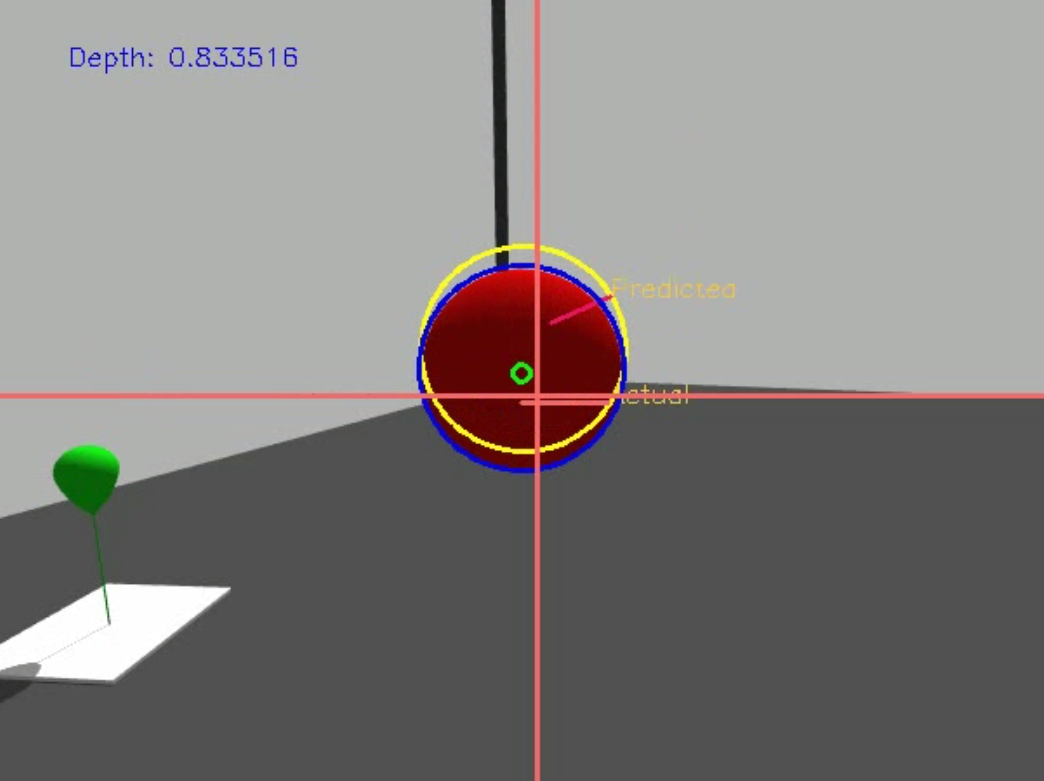}
	\caption{}
\end{subfigure}
    \begin{subfigure}{0.45\columnwidth}
	\centering
 \includegraphics[scale=0.375]{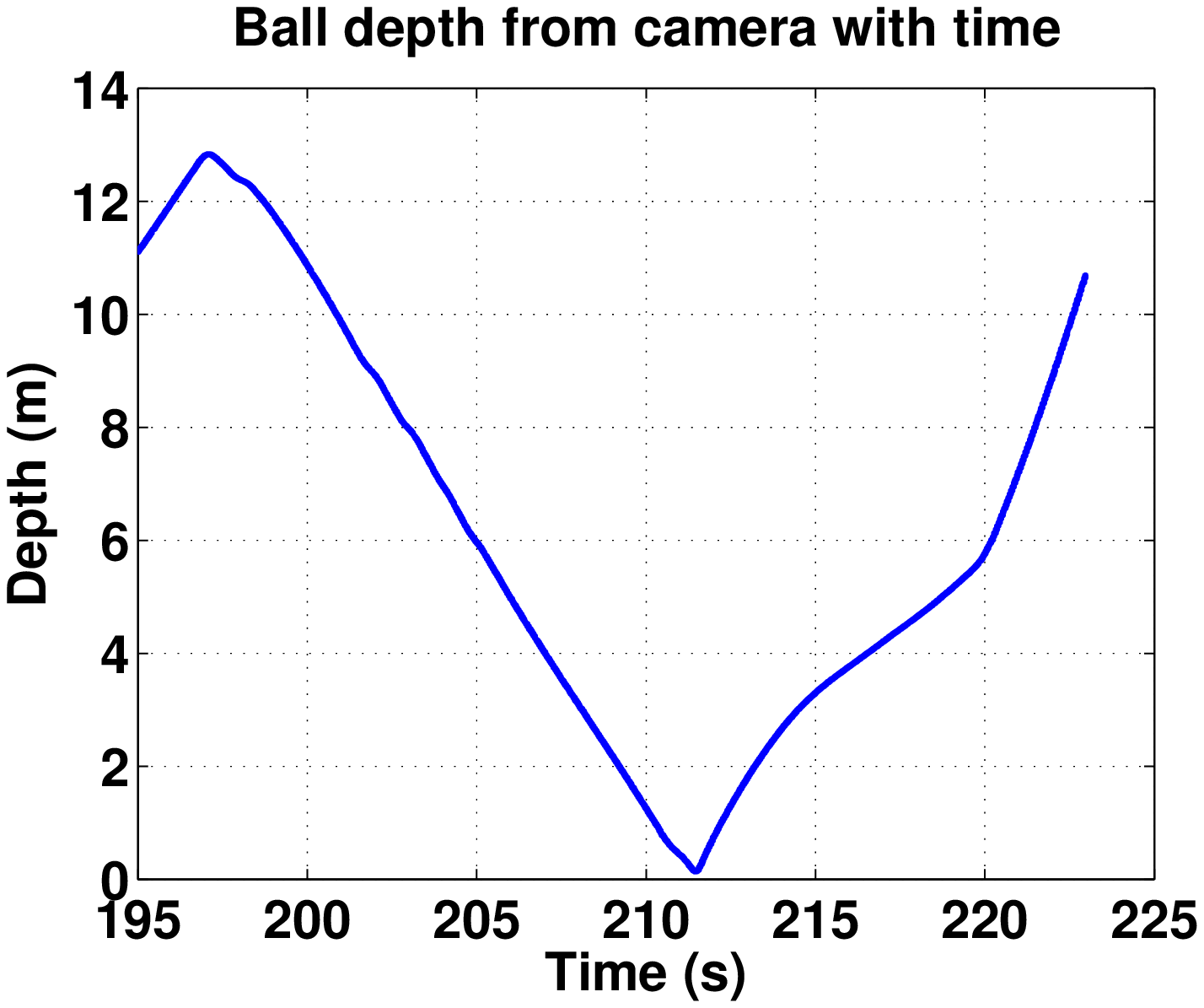}
	\caption{}
\end{subfigure}
    \begin{subfigure}{0.45\columnwidth}
	\centering
 \includegraphics[scale=0.22]{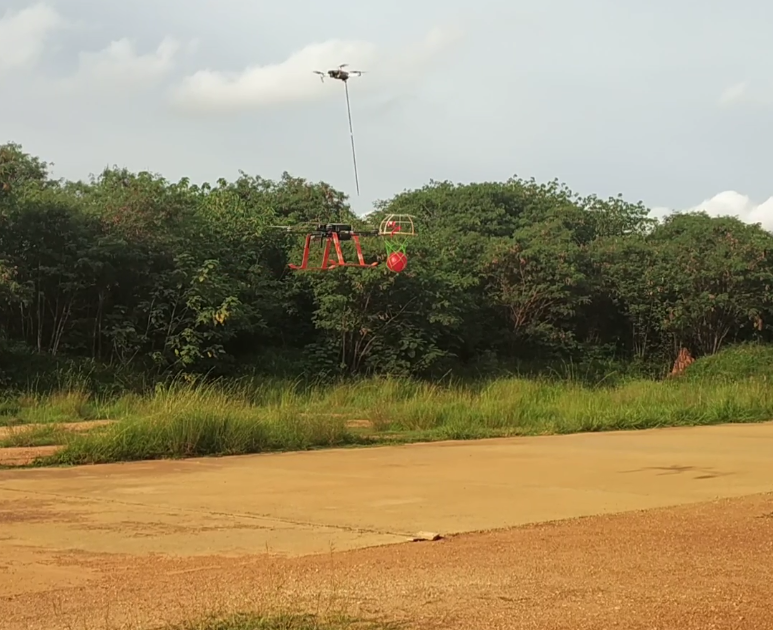}
	\caption{}
\end{subfigure}
   \caption{(a)-(b) Ball capture in Gazebo environment (c) Ball depth profile from grabbing drone's camera (d)UAV grabbing a ball suspended from target drone in realistic scenario}
    \label{f9}
\end{figure}

The software set up for the same is first developed in Robot Operating System (ROS) architecture and is visualised in Gazebo environment. For this, IRIS drone models are imported (Fig. \ref{f8}). 
The tracking and grabbing modules are tested in the simulation environment before porting to the real hardware. Fig. \ref{f8}(a) shows the target being tracked by own UAV. Fig. \ref{f9}(a)-(b) are snap shots of grabbing from Gazebo simulation environment, where, the UAV end effector camera information is utilised to align with the incoming target ball. The distance of ball from the gripper camera decreases as the ball reaches near the camera. The ball capture occurs at 211 s, as can be seen in Fig. \ref{f9}(c). The videos of the same can be found out in \footnote{\href{https://indianinstituteofscience-my.sharepoint.com/:f:/g/personal/limatony_iisc_ac_in/EnFgJZVt4r1GncZg9TA_7Z0BNtToa6KbSYqNPh7ge4oRnQ?e=zaDD2I}{Gazebo sim}}. 

\begin{figure}
    \centering
        \begin{subfigure}{0.45\textwidth}
\includegraphics[scale=0.5]{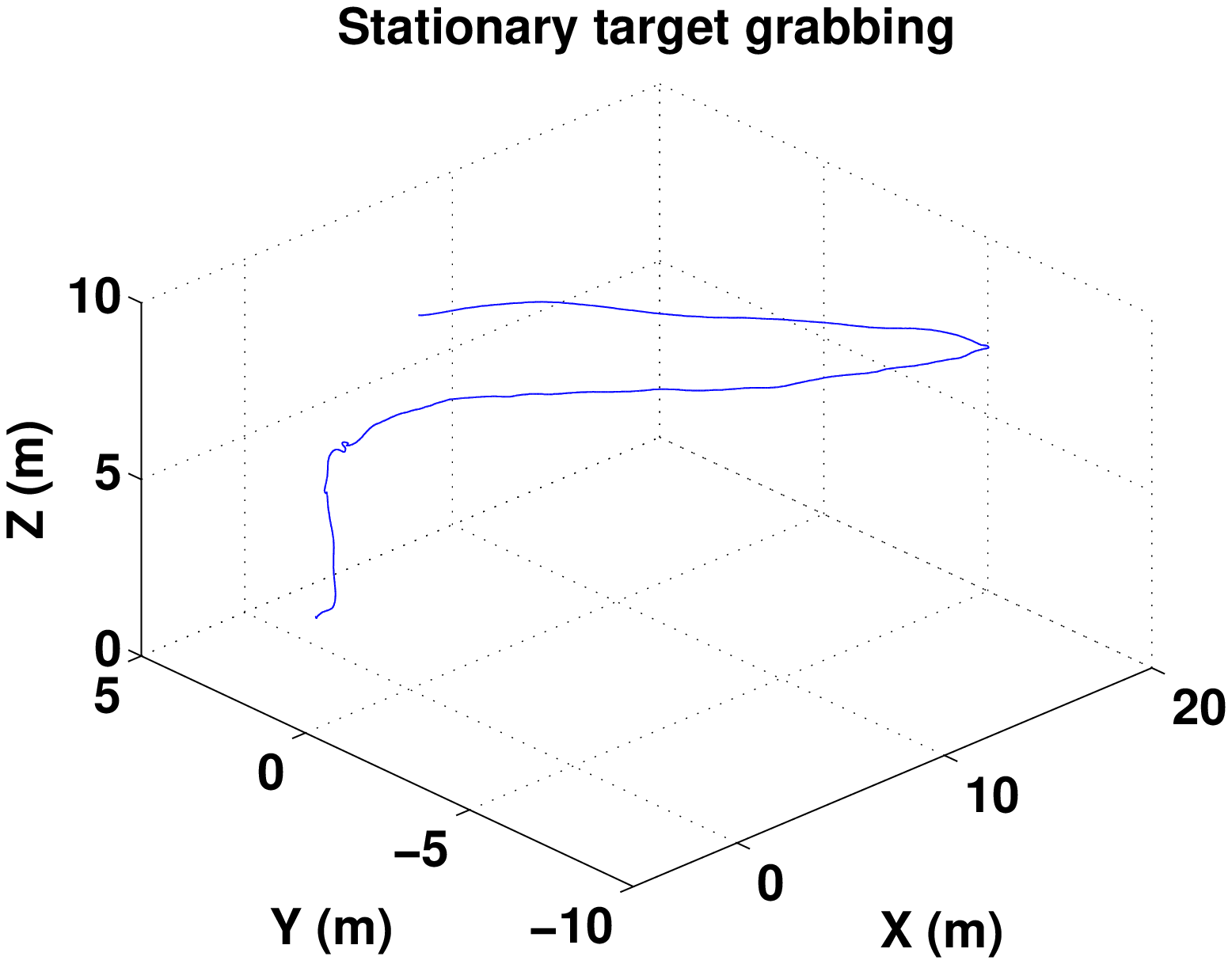} 
\caption{}
\end{subfigure}
    \begin{subfigure}{0.45\textwidth}
\includegraphics[scale=0.5]{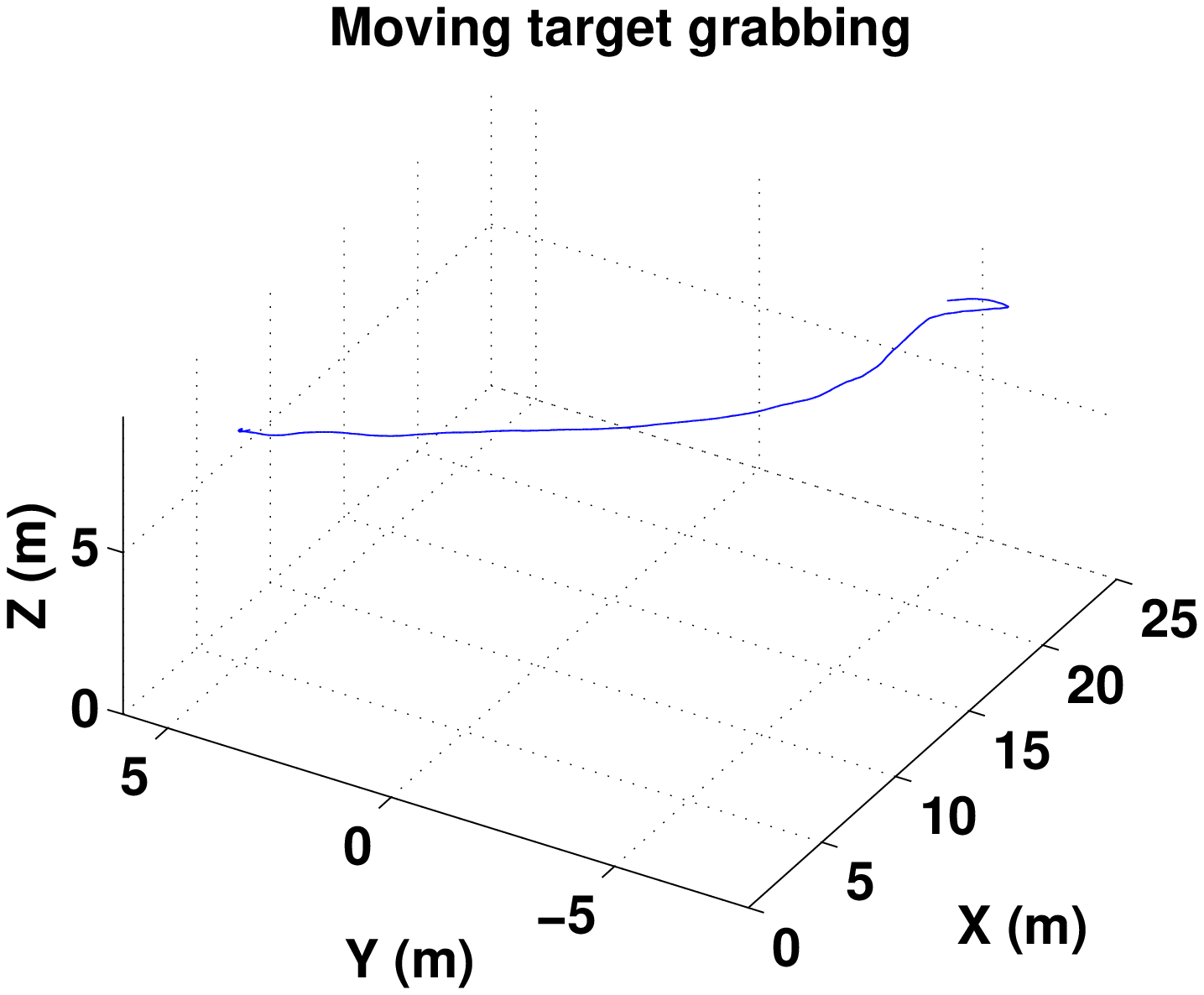} 
\caption{}
\end{subfigure}
    \caption{UAV trajectory for (a) static ball capture (b) moving ball capture}
    \label{f9a}
\end{figure}

The same is ported on to the hardware platform. GA3 drone uses visual feedback from the manipulator camera to achieve tracking. Here, the target drone with a red ball moves in figure of eight. Own UAV detects the target and does perfect tracking (snap shot of the same in Fig. \ref{f8}(b)). Video of the same can be seen at \footnote{\href{https://indianinstituteofscience-my.sharepoint.com/:v:/g/personal/limatony_iisc_ac_in/EQR1XDVwlZJPmb7aB49Vsf8BPe9NdoQFvoozuhgNW_C7Vg?e=8CNnQV}{Target tracking}}. Following this, a static target grabbing is tested. The target drone with a red ball suspended by a rod is considered the stationary target. The vehicle achieves successful grabbing\footnote{\href{https://indianinstituteofscience-my.sharepoint.com/:v:/g/personal/limatony_iisc_ac_in/EYeAjVB1LiNBvGV6UsQK4pwB_PBWpLYdyQfOP2WAj8HArw?e=ScHCGg}{Stationary target grabbing}}, as shown in Fig. \ref{f9}.
Next step of implementation is grabbing of moving ball.The same set of vehicles are used except that the target drone is slowly moving in a straight line\footnote{\href{https://indianinstituteofscience-my.sharepoint.com/:v:/g/personal/limatony_iisc_ac_in/Ec1_DebV9clErDHsQX8S3EMBWNLaH08Ami5VmS51zKzBpA?e=nXxa24}{Moving target grabbing}}. The trajectory of grabber drone in each of these cases are in Fig. \ref{f9a}. Fig. \ref{f9a}(a) shows the vehicle taking off, approaching the target and grabbing. Fig. \ref{f9a}(b) shows the UAV chasing the target and detaching the ball. 

\begin{figure}
    \centering
    \includegraphics[scale=0.325]{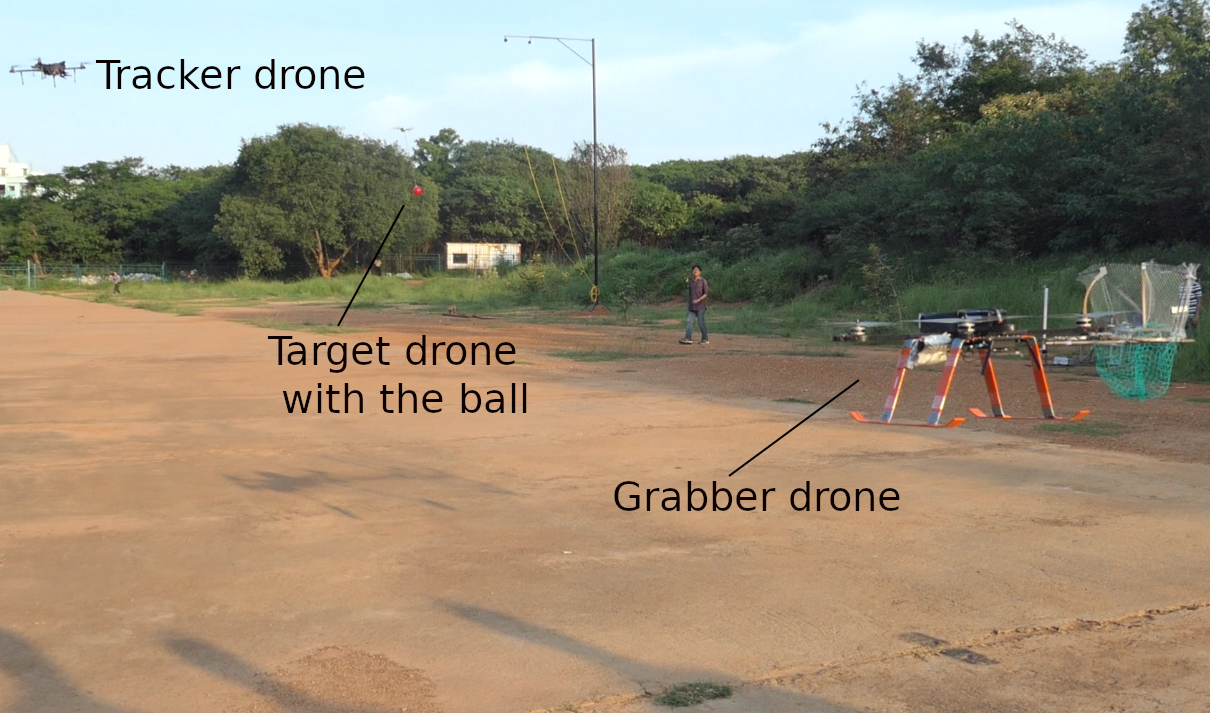}
    \caption{UAV grabbing a ball suspended from target drone}
    \label{f10}
\end{figure}

Extending further, the next experiment is collaborative grabbing using two drones. Tracking drone takes off to search for target. Upon detection, the target location is communicated to grabber drone. The second drone takes off towards the communicated location and continues until the target ball is in FOV. After detection, the grabber UAV proceeds towards the ball to successfully grab and land\footnote{\href{https://indianinstituteofscience-my.sharepoint.com/:v:/g/personal/limatony_iisc_ac_in/EbPDqHhzuJ9KiEDCxXh8YoYB1sT_k8AtmA7uWbH84KZe9w?e=PIqJgg}{Collaborative grabbing}}. A snap shot of the same is shown in Fig. \ref{f10}.

The grabbing of stationary ball is achieved with a success rate of 8/10 while the capture of moving target has a success rate of 7/10. The major factors that govern the failure in this case are loss of target from FOV or external disturbances like wind which pushes the target ball away at the terminal phase of grabbing. Introducing collaborative grabbing improves the ball capture rate to 9/10 as a consistent feedback is obtained from a third eye thus improving the grabber chances.
\section{Conclusions}\label{S5}
This work details the  hardware implementation of collaborative  aerial target capture using multiple UAVs. Vision based target detection and guidance is developed and validated through simulation in ROS frame work. An exclusive manipulator mechanism is designed which improves the efficiency of the mission and is prototyped for hardware implementation. Individual modules of tracking and grabbing are tested in Gazebo environment which is further ported into real time system. The results achieved is improved using collaborative capture where a third UAV keeps the target in its FOV to aid the grabbing UAV. This improves the efficiency of the entire pipeline.  Future work involves optimum task allocation among collaborative UAVs for capturing of high speed and maneuverable target.

\begin{acknowledgements}
	We would like to acknowledge the Robert Bosch Center for Cyber Physical Systems, Indian Institute of Science, Bangalore, and Khalifa University, Abu Dhabi, for partial financial support. We would also like to thank fellow team members from IISc for their invaluable contributions towards this competition. 
\end{acknowledgements}

\end{document}